\documentclass[11pt,letterpaper]{article}

\usepackage{xcolor}
\definecolor{navy}{rgb}{0,0.1,0.4}
\definecolor{navy-cap}{rgb}{0,0.1,0.5}
\definecolor{navy-ref}{rgb}{0,0.3,1}
\definecolor{lgray}{gray}{0.90}
\definecolor{dkgreen}{rgb}{0,0.6,0}
\definecolor{gray}{rgb}{0.5,0.5,0.5}
\definecolor{mauve}{rgb}{0.58,0,0.82}

\usepackage{sectsty}
\sectionfont{\fontsize{12}{15}\selectfont \color{navy}}
\subsectionfont{\fontsize{12}{15}\selectfont \color{navy}}
\subsubsectionfont{\fontsize{12}{15}\selectfont \color{navy}}

\usepackage[margin = 1 in]{geometry}

\usepackage{times}
\usepackage[font={color=navy-cap, small}, labelfont={bf}, labelsep=space, justification=justified, skip=0pt]{caption}
\usepackage[bottom, hang,flushmargin]{footmisc} 

\usepackage{titlesec}
\usepackage{tikz}
\titlespacing{\section}{0pt}{0.75em}{0.25em}
\titlespacing{\subsection}{0pt}{0.5em}{0.25em}
\titlespacing{\subsubsection}{0pt}{0.5em}{0.25em}

\usepackage{array,longtable}
\usepackage{xltabular}
\usepackage{multirow}
\usepackage{pifont}
\usepackage[font={color=navy-cap, footnotesize}, labelfont={bf}, labelsep=space, justification=justified, skip=3pt]{caption}
\usepackage{subcaption}
\usepackage{booktabs} 
\usepackage{graphicx}
\usepackage{wrapfig}
\usepackage{enumitem}
\usepackage{hyperref}
\hypersetup{colorlinks = true, linkcolor = navy-ref}
\usepackage{textcomp}
\usepackage{url}
\usepackage{amssymb}
\usepackage{bm}
\usepackage{amsmath}

\usepackage{dirtytalk}
\usepackage{pdfpages} 
\usepackage{soul} 
\usepackage{fancyhdr}
\usetikzlibrary{patterns}
\usepackage{pgfplots}
\usepackage{floatrow}
\usepackage{setspace}
\usepackage{bbm}

\usepackage[
backend=biber,
style=numeric-comp,
sorting=none,
maxnames = 10
]{biblatex}
\usepackage[nameinlink]{cleveref} 

\usepackage{colortbl}
\usepackage{hhline}

\newcommand{\cmt}[1]{} 
\newcommand\brackets[1]{\mathopen{}\left[#1\right]\mathclose{}}
\newcommand\braces[1]{\mathopen{}\left \{ #1\right \} \mathclose{}}
\newcommand\parens[1]{\mathopen{}\left(#1\right)\mathclose{}}
\newcommand\bars[1]{\left|#1\right|}

\newcommand{\betab}{\boldsymbol{\beta}}
\newcommand{\bb}{\boldsymbol{b}}
\newcommand{\Cb}{\boldsymbol{C}}
\newcommand{\Cblambda}{\boldsymbol{C_\lambda}}

\newcommand{\eg}{e.g., }

\newcommand{\gp}{\texttt{GP+}}
\newcommand{\analyticI}{\texttt{Analytic I}}
\newcommand{\analyticII}{\texttt{Analytic II}}

\newcommand{\hartmann}{\texttt{Hartmann}}
\newcommand{\volcano}{\texttt{Volcano}}
\newcommand{\elevators}{\texttt{Elevators}}
\newcommand{\pumadyn}{\texttt{Pumadyn}}
\newcommand{\rom}{\texttt{ROM}}

\newcommand{\omegab}{\boldsymbol{\omega}}

\newcommand{\ie}{i.e., }

\newcommand{\mb}{\boldsymbol{m}}

\newcommand{\mathD}{\mathcal{D}}

\newcommand{\mathL}{\mathcal{L}}
\newcommand{\mathN}{\mathcal{N}}

\newcommand{\mathX}{\mathcal{X}}

\newcommand{\mathZ}{\mathcal{Z}}

\newcommand{\real}{\mathbb{R}}
\newcommand{\thetab}{\boldsymbol{\theta}}

\newcommand{\short}{SEEK}
\newcommand{\shortSpace}{SEEK }

\newcommand{\xb}{\boldsymbol{x}}

\newcommand{\wb}{\boldsymbol{w}}
\newcommand{\Xb}{\boldsymbol{X}}
\newcommand{\ab}{\boldsymbol{a}}

\newcommand{\yb}{\boldsymbol{y}}

\usepackage[ruled,vlined]{algorithm2e}

\usepackage{bbm}
\usepackage{xfrac}
\usepackage{enumitem}

\usepackage{amsthm}
\newtheorem{definition}{Definition}

\newtheorem{theorem}{Theorem}
\usepackage{enumitem}
\usepackage{cancel}
\usepackage{mathtools}

\usepackage{placeins}

\newtheorem{condition}{Condition}
\newtheorem{property}{Property}

    \makeatletter
\def\@fnsymbol#1{\ensuremath{\ifcase#1\or \dagger\or *\or \ddagger\or
   \mathsection\or \mathparagraph\or \|\or **\or \dagger\dagger
   \or \ddagger\ddagger \else\@ctrerr\fi}}
    \makeatother

\title{SEEK: Self-adaptive Explainable Kernel For Nonstationary Gaussian Processes}
\date{\vspace{-5ex}}
\usepackage{authblk}
\author[1]{Nima Negarandeh$^\dagger$}
\author[1]{Carlos Mora\thanks{Equal Contribution}}
\author[1]{Ramin Bostanabad\thanks{\noindent Corresponding Author: Raminb@uci.edu}}
\affil[1]{University of California, Irvine, CA, United States of America}

\begin{document}
    \pagenumbering{arabic}
    \maketitle
    \sloppy 
    \noindent \textcolor{navy}{\textbf{Abstract}}

Gaussian processes (GPs) are powerful probabilistic models that define flexible priors over functions, offering strong interpretability and uncertainty quantification. However, GP models often rely on simple, stationary kernels which can lead to suboptimal predictions and miscalibrated uncertainty estimates, especially in nonstationary real-world applications. In this paper, we introduce \short, a novel class of learnable kernels to model complex, nonstationary functions via GPs. Inspired by artificial neurons, \shortSpace is derived from first principles to ensure symmetry and positive semi-definiteness, key properties of valid kernels. The proposed method achieves flexible and adaptive nonstationarity by learning a mapping from a set of base kernels. Compared to existing techniques, our approach is more interpretable and much less prone to overfitting.
We conduct comprehensive sensitivity analyses and comparative studies to demonstrate that our approach is not only robust to many of its design choices, but also outperforms existing stationary/nonstationary kernels in both mean prediction accuracy and uncertainty quantification.

\noindent \textcolor{navy}{\textbf{Keywords:}} 
Nonstationary Kernels; Gaussian Processes; Neural Networks. 

    \section{Introduction} \label{sec: intro}

Gaussian processes (GPs) are a class of powerful yet interpretable semi-parametric Bayesian models that define flexible prior distributions over functions \cite{williams2006gaussian}. Their natural ability to quantify uncertainty has made them a valuable tool for researchers and practitioners across various disciplines. 

A GP, typically denoted as $GP \parens{m\parens{\xb;\thetab}, c\parens{\xb,\xb^{\prime};\betab}}$, is fully characterized by its mean function $m\parens{\xb;\thetab}$ and covariance function or kernel $c\parens{\xb,\xb^{\prime};\betab}$ with parameters $\thetab$ and $\betab$, respectively. Many options exist for choosing the mean and covariance functions but zero-mean GPs with stationary kernels, \ie kernels that depend only on the relative distance between data points rather than their absolute positions, have been the standard choice over the past decades \cite{pilario2019review}. While not always optimal, stationary kernels are widely used due to their general effectiveness, simplicity, and computational efficiency.

GPs with stationary kernels and simple mean functions can provide suboptimal performance in terms of predictive mean and, more importantly, variance \cite{noack2023mathematical, chen2024adaptive}. This limitation is especially concerning for tasks where reliable uncertainty quantification (UQ) is essential, such as Bayesian optimization (BO) \cite{frazier2018tutorial, shahriari2015taking}. In such cases, miscalibrated uncertainty estimates can lead to inefficient exploration and suboptimal decision-making. To address these limitations, substantial efforts have been made to enhance GP emulation capabilities to model nonstationary functions, either through flexible mean functions \cite{iwata2017improving}, nonstationary kernels \cite{gibbs1998bayesian,paciorek2003nonstationary}, or both \cite{ba2012composite}. While flexible mean functions can enhance a GP's predictive accuracy in nonstationary problems, they fail to fully correct the predictive variance. Consequently, developing nonstationary kernels remains a more effective approach for improving both mean and uncertainty predictions. 

Unlike stationary kernels that can be formulated as $c(\xb,\xb^{\prime};\betab) = c\parens{\xb-\xb^{\prime};\betab}$, nonstationary kernels allow the relationship to depend on the \textit{absolute} positions of the data points, \ie $c(\xb,\xb^{\prime};\betab) \neq c\parens{\xb-\xb^{\prime};\betab}$. This makes nonstationary kernels much more flexible to capture fine-grained variations across the entire domain. However, effectively designing them remains an open research challenge, as it is unclear how to best provide this flexibility without introducing new issues. While one might assume that GPs are inherently safeguarded against overparameterization due to their probabilistic nature, research has shown that naively increasing kernel/mean complexity can lead to severe optimization difficulties and overfitting \cite{ober2021promises, noack2024unifying, van2021feature}.


Numerous techniques have been proposed to build nonstationary kernels and we broadly categorize them into three types \cite{chen2024adaptive}: (1) input-dependent lengthscales \cite{gibbs1998bayesian, paciorek2003nonstationary, heinonen2016non}, (2) input warping \cite{wilson2016deep, snoek2014input, tompkins2020sparse}, and (3) mixture of GP experts \cite{yuksel2012twenty}. 

The input-dependent lengthscale approach modifies the kernel function by allowing the lengthscale parameters to depend on the inputs. Notable examples include the Gibbs kernel \cite{gibbs1998bayesian}, which explicitly incorporates input-dependent smoothness by modeling the lengthscale function using either a GP or a neural network (NN) \cite{paciorek2003nonstationary, plagemann2008nonstationary, remes2018neural}. However, the lack of structural constraints on the lengthscale function often poses overfitting and non-identifiability issues \cite{tompkins2020sparse, chen2024adaptive}, since many lengthscale functions can yield similar likelihood values. 

The input warping approach transforms the input space via a mapping before applying the kernel. These transformations can be learned \cite{wilson2016deep, snoek2014input} or predefined using a set of basis functions \cite{gibbs1998bayesian, xiong2007non}. A well-known recent example is the deep kernel \cite{wilson2016deep}, which employs NNs to learn an expressive input transformation before applying the kernel. 
The concept of input warping dates back several decades, originating from the characterization of stationary reducible and locally stationary reducible kernels \cite{genton2001classes, genton2004time, sampson1992nonparametric} where the key idea is to learn a feature space where stationarity or local stationarity holds. A key limitation of this approach is that the learned mapping must be bijective to ensure a valid transformation \cite{genton2001classes}. For example, \cite{snoek2014input} models the mapping using the Beta cumulative distribution function, which can represent a broad class of bijective functions while having few hyperparameters.

Lastly, the mixture of experts approach partitions the input space into different regions and assigns distinct GP models to each, allowing for locally adaptive behavior \cite{yuksel2012twenty, rasmussen2001infinite, trapp2020deep}. Typically, these models employ gating functions to weight contributions from different experts and ensure smooth transitions across regions \cite{rasmussen2001infinite, meeds2005alternative}. However, scaling this approach to high-dimensional problems is challenging, as the number of experts must increase dramatically with the input dimensionality. 

A more recent approach that lies at the intersection of the above three categories is the attentive kernel (AK) \cite{chen2024adaptive}. AK is designed to mitigate the training challenges of nonstationary kernels by introducing similarity attention scores to weight a predefined set of basis kernels and visibility attention scores to mask out data across sharp transitions. This approach essentially selects relevant subsets of data in prediction and has demonstrated improved mean and uncertainty estimates while reducing overfitting compared to conventional nonstationary kernels, particularly in 2D and 3D robotics applications. AK fails to scale to high dimensional inputs as the number of predefined basis kernels must increase substantially with input dimensionality.

Our proposed method introduces a new category for learning nonstationary kernels. We provide a structured way to construct expressive kernels that scale well and can capture complex nonstationary patterns while being robust to overfitting. In essence, our idea is to build a nonstationary kernel via a self-adaptive composition of a set of base kernels. The self-adaptivity nature of our kernel underlies its nonstationarity feature and is achieved by integrating learnable and input-dependent weights with the base kernels such that the resulting composition not only remains interpretable, but also is guaranteed to be a valid kernel. We call our kernel SElf-adaptive Explainable Kernel or \short. 

The remainder of the paper is structured as follows. In Section \ref{sec method}, we introduce \shortSpace and then present a comprehensive evaluation of its performance on multiple benchmark problems in Section \ref{sec results}. We summarize our contributions and discuss potential directions for future research in Section \ref{sec conclusion}.
    \section{Methods} \label{sec method}
We briefly review GPs in Section \ref{subsec method gp}. Then, we discuss the conditions that a valid covariance function should satisfy and the main closure properties of kernels in Sections \ref{subsec valid kernel} and \ref{subsec closure properties}, respectively. These discussions set the stage for introducing \shortSpace in Section \ref{subsec method proposed framework}. 

\subsection{Gaussian Processes (GPs)}\label{subsec method gp}
A GP is a stochastic process whose samples follow a multivariate normal distribution. 
In the context of regression problems, we consider a training dataset $\mathD \coloneqq \braces{(\xb_i,y_i)}_{i=1}^N$, and denote the collection of inputs by $\Xb = \brackets{\xb_1,\dots,\xb_N}^T$, where $\xb_i \in \mathX \subset \real^P$, with corresponding outputs $\yb = \brackets{y_1,\dots,y_N}^T$, where $y_i \in \real$. We assume that the samples are generated according to the model\footnote{The GP framework can be easily extended to multi-dimensional outputs but for simplicity we present the formulation for the single-output case.}: 
\begin{equation} 
    y(\xb) = f(\xb) + \epsilon,
    \label{eq general regression model}
\end{equation}
where $f(\xb)$ is the unknown latent function and $\epsilon \sim \mathN(0, \lambda^2)$ is the independent Gaussian noise with variance $\lambda^2$. Under this framework, we place a GP prior on $f(\xb)$, \ie $f\parens{\xb}\sim GP \parens{m\parens{\xb;\thetab}, c\parens{\xb,\xb^\prime;\betab}}$.
Hereafter, we omit the dependencies of $m(\xb)$ on $\thetab$ and $c\parens{\xb,\xb^\prime}$ on $\betab$ to improve readability.

A key property of GPs is that they are closed under Bayesian conditioning. This implies that the posterior distribution of $f(\xb)$ conditioned on the observed data $\mathD$ remains a GP, \ie $f(\xb)|\mathD \sim GP \parens{\bar{m}\parens{\xb}, \bar{c}\parens{\xb,\xb^\prime}}$. Therefore, the posterior mean and variance at any unseen input $\xb^*$ have the following closed-form expressions:
\begin{subequations} 
    \begin{equation} 
        \bar{m}\parens{\xb^*} = m\parens{\xb^*} + c\parens{\xb^*,\Xb} \Cblambda^{-1}\parens{\yb - \mb},
    \label{eq posterior mean}
    \end{equation}
    \begin{equation}
        \bar{c}\parens{\xb^*,\xb^*} = c\parens{\xb^*,\xb^*} - c\parens{\xb^*,\Xb}\Cblambda^{-1}c\parens{\Xb,\xb^*},
    \label{eq posterior cov}
    \end{equation}
    \label{eq posterior}
\end{subequations}    
where $\Cblambda = \Cb + \lambda^2I$ is the $N\times N$ covariance matrix with $\Cb = c\parens{\Xb,\Xb}$, and $\mb = m(\Xb)$. For the purpose of this paper, we only consider zero-mean functions, \ie $m(\xb)=0$, and focus on the kernel for modeling nonstationary functions. 

Although one could use the posterior equations from Eq. \ref{eq posterior} directly for predictions without estimating the kernel parameters $\betab$ and noise variance $\lambda^2$, it is common practice to learn them via maximum likelihood estimation (MLE), that is: 
\begin{equation}
        \brackets{\hat{\betab}, \hat{\lambda^2}} = \underset{\betab, \lambda^2}{\operatorname{argmin}} \ \mathL = \underset{\betab, \lambda^2}{\operatorname{argmin}}  \ \frac{1}{2} \log{\bars{\Cblambda}} + \frac{1}{2} \yb^T \Cblambda^{-1} \yb .
\end{equation}


Typical choices for $c\parens{\xb,\xb^{\prime}}$ are the stationary Gaussian and Matérn kernels: 
\begin{subequations} 
    \begin{equation} 
        c(\xb, \xb^{\prime}) =
        \exp \parens{-d^2},
    \label{eq Gaussian kernel}
    \end{equation}
    \begin{equation}
        c(\xb, \xb^{\prime}) = \frac{2^{1-\nu}}{\Gamma(\nu)} 
        \parens{\sqrt{2\nu} d}^{\nu}
        K_{\nu} \parens{\sqrt{2\nu} d}.
    \label{eq Matern kernel}
    \end{equation}
    \label{eq common kernels}
\end{subequations}   
In both cases, $d = \sqrt{\parens{\xb-\xb^\prime}^T \text{diag}\parens{10^{\omegab}} \parens{\xb-\xb^\prime}}$, where $\omegab \in \real^P$ is the vector of lengthscale parameters. In the Matérn kernel, $K_{\nu}$ is the modified Bessel function of the second kind, and $\Gamma$ is the gamma function. It is common practice to fix $\nu$ to half-integer values $\frac{1}{2}$,$\frac{3}{2}$ or $\frac{5}{2}$ to simplify the $\Gamma$ and $K_{\nu}$ functions, leading to simpler closed-form expressions of Matérn kernel that substantially reduce computational costs. 

Stationary covariance functions, such as those in Equations \ref{eq Gaussian kernel} and \ref{eq Matern kernel}, introduce the inductive bias that nearby input points yield correlated outputs. However, an extensive class of functions can serve as valid GP kernels provided that they satisfy certain conditions, which we review in the following sections.

\subsection{Validity of Kernels for GPs}\label{subsec valid kernel}
To serve as a valid kernel for GPs, a function must satisfy two necessary conditions:

\begin{condition}[Symmetry]\label{def symmetry}
A function $c:\mathX \times \mathX \rightarrow \real$ is symmetric if
\[
  c(\xb, \xb') = c(\xb', \xb),
  \quad \forall \xb, \xb'.
\]
\end{condition}

\begin{condition}[Positive semi-definiteness]\label{def psd}
A function $c:\mathX \times \mathX \rightarrow \real$ is positive-semidefinite (PSD) if, for any finite set of points 
\(\{\xb_1, \xb_2, \dots, \xb_n\}\subset \mathcal{X}\), 
the resulting covariance matrix $\Cb = \bigl[c(\xb_i, \xb_j)\bigr]_{i,j=1}^n$ satisfies 
\[
  \ab^\top \Cb \ab \geq 0, \quad \forall \ab \in \real^n, \ab \neq \boldsymbol{0}.
\]
\end{condition}

Conditions \ref{def symmetry} and \ref{def psd} ensure 
that $c(\xb, \xb')$ is symmetric and PSD, making it a valid kernel for GPs.

\subsection{Some Closure Properties on Kernels}\label{subsec closure properties}
We now present four fundamental closure properties of kernels. These properties form the theoretical foundation of \short. 

Let $c_1:\mathX \times \mathX \rightarrow \real$, $c_2:\mathX \times \mathX \rightarrow \real$ and $c_3:\mathZ \times \mathZ \rightarrow \real$ be valid kernels, with $\psi:\mathX \rightarrow \mathZ \subset \real^Z$. Then, the following functions are also valid kernels:

\begin{property}[Scaling]\label{def scaling}  
    \[
    c(\xb,\xb') = \alpha\, c_1(\xb,\xb'), \quad \forall \alpha \geq 0.
    \]
\end{property}

\begin{property}[Addition]\label{def addition}
    \[
    c(\xb,\xb') = \alpha_1 c_1(\xb,\xb') + \alpha_2 c_2(\xb,\xb'), \quad \forall \alpha_1,\alpha_2 \ge 0.
    \]
\end{property}

\begin{property}[Product]\label{def product}
    \[
    c(\xb,\xb') = c_1(\xb,\xb')c_2(\xb,\xb').
    \]
\end{property}

\begin{property}[Warping]\label{def warping}
    \[
    c(\xb,\xb') = c_3(\psi(\xb),\psi(\xb')).
    \]
\end{property}

Building upon Properties \ref{def scaling}$-$\ref{def product}, it follows that if $f$ is a polynomial with non-negative coefficients, then $f(c(\xb,\xb^{\prime}))$ is also a valid kernel. Theorem \ref{thm:hadamard_powers} further generalizes kernel validity by establishing conditions under which polynomial transformations and related analytic expansions preserve kernel validity.



\begin{theorem}[Kernel Validity under Analytic Transformations {\cite[Theorem 7.5.9]{horn2012matrix}}]
\label{thm:hadamard_powers}
Let \(Z = [z_{ij}] \in \real^{n \times n}\) be positive semidefinite.

\begin{enumerate}
\item The Hadamard powers \(Z^{(k)} = [z_{ij}^k]\) are positive semidefinite for all \(k = 1,2,\ldots\); they are positive definite if \(Z\) is positive definite.

\item Let \(f(z) = a_0 + a_1 z + a_2 z^2 + \cdots\) be an analytic function with nonnegative coefficients and radius of convergence \(R > 0\). Then \(\bigl[f(z_{ij})\bigr]\) is positive semidefinite if \(\lvert z_{ij}\rvert < R\) for all \(i,j\in\{1,\dots,n\}\); it is positive definite if, in addition, \(Z\) is positive definite and \(a_i > 0\) for some \(i\in\{1,2,\ldots\}\).

\item The Hadamard exponential matrix \(\bigl[e^{z_{ij}}\bigr]\) is positive semidefinite; it is positive definite if and only if no two rows of \(Z\) are identical.
\end{enumerate}
\end{theorem}

From this theorem, it follows that any analytic function $f\parens{\cdot}$ which can be expressed (or approximated) as a power series (e.g., a Taylor series expansion) with all non-negative coefficients, preserves Condition \ref{def psd} (PSD) within its radius of convergence. Moreover, since such a function maintains symmetry (Condition \ref{def symmetry}), it also preserves the validity of the kernel.

This result can be leveraged to show that the Gaussian kernel is indeed a valid kernel, or more generally, that the exponential transformation $\exp{\parens{c(\xb,\xb^\prime)}}$ preserves kernel validity. This reasoning naturally extends to other transformations, including $\sinh{\parens{c(\xb,\xb^\prime)}}$ and $\cosh{\parens{c(\xb,\xb^\prime)}}$. 

Regarding Property \ref{def warping}, we note that it underlies input warping techniques such as deep kernels \cite{wilson2016deep} where a learned feature mapping $\psi(\xb)$ is used to transform the input space before applying a base kernel.


These closure properties provide a principled and structured framework for systematically constructing new kernels from existing ones while preserving the fundamental requirements of symmetry and positive semi-definiteness, as outlined in Conditions \ref{def symmetry} and \ref{def psd}. In Section \ref{subsec method proposed framework}, we leverage Properties \ref{def scaling}$-$\ref{def warping} and Theorem \ref{thm:hadamard_powers} to introduce \short.

\subsection{\short: Our Proposed Kernel} \label{subsec method proposed framework}

\shortSpace is inspired by the architecture of a single artificial neuron in NNs \cite{mcculloch1943logical}. In a typical neuron, each feature is multiplied by a weight, a bias term is added, and the result is passed through an activation function to capture complex patterns. Following this intuition, we design \shortSpace by applying an appropriate nonlinear activation function to a weighted sum of base kernels and an added bias term, ensuring validity through kernel closure properties (Properties \ref{def scaling}$-$\ref{def warping}) and Theorem \ref{thm:hadamard_powers}. In \short, the weights and the bias are learnable functions of the inputs (\eg parameterized  by NNs), making the overall kernel nonstationary. Definition~\ref{def kernelX} lays out the mathematical form of \shortSpace which is formulated in Equation~\eqref{eq kernelX}.

\begin{figure*}[!t] 
    \centering
    \includegraphics[width=\linewidth]{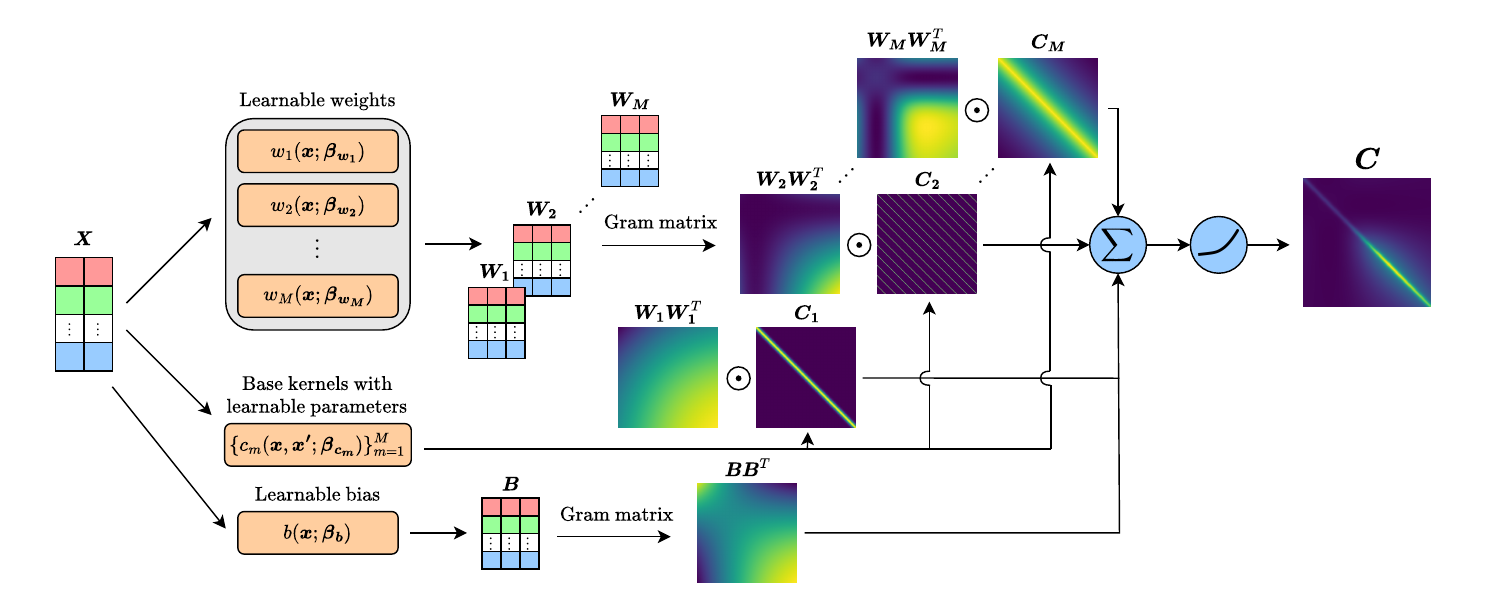}
    \caption{Schematic illustration of our kernel: \short~has a set of weighted base kernels with learnable hyperparameters where the weights are learnable functions. These weighted kernels, along with a learnable bias term, are summed and then passed through an appropriate activation function to produce the final nonstationary covariance function.}
    \label{fig workflow}
\end{figure*}

\begin{definition}[\shortSpace Kernel]
Let $\xb,\xb^{\prime} \in \real^P$ be two input points, and consider vector-valued functions $\wb_m(\xb; \betab_{\wb_m}): \real^P \rightarrow \real^{W_m}$ and $\bb(\xb; \betab_b): \real^P \rightarrow \real^{B}$, parameterized via $\betab_{\wb_m}$ and $\betab_b$, respectively.\footnote{Unless otherwise specified, all vectors are treated as column vectors.} Given a set of $M$ base kernels $\braces{c_m\parens{\xb,\xb^{\prime}; \betab_{c_m}}}_{m=1}^M$ and an appropriate activation function $\phi:\real \rightarrow \real$, we define \shortSpace as:
    \begin{equation}
        c(\xb, \xb^{\prime}) = \phi\parens{z\parens{\xb,\xb^\prime}},
        \label{eq kernelX}
    \end{equation}
    with the pre-activation $z(\xb,\xb^\prime)$ defined as:  
\begin{equation}
    z(\xb,\xb^\prime) = \sum_{m=1}^M \wb_m^{\top}(\xb) \wb_m(\xb^{\prime}) c_m(\xb, \xb^{\prime}) + \bb^\top(\xb) \bb(\xb^{\prime}).
    \label{eq preactivation}
\end{equation}
    \label{def kernelX}
\end{definition}
\vspace{-5mm}
\noindent where we have dropped the dependence of the functions in Equations \ref{eq kernelX} and \ref{eq preactivation} on their parameters to improve readability. 

Figure~\ref{fig workflow} illustrates the overall workflow of \short, where the base kernels $c_m(\xb, \xb^{\prime})$ are weighted, combined, and then passed through an activation function to produce the final kernel output. This formulation enables \shortSpace to flexibly capture complex, nonlinear relationships by modulating the contributions of multiple base kernels adaptively across the entire input space. We now discuss three important features of \short.

First, the weight functions $\wb_m(\xb)$ and the bias function $\bb(\xb)$ can be modeled using different function approximators, each offering varying levels of expressiveness and complexity. Potential choices include polynomials, NNs, or GPs.

The second feature is the flexibility in the choice of base kernels $\braces{c_m\parens{\xb,\xb^{\prime}}}_{m=1}^M$, which can include any valid stationary (\eg Gaussian kernel) and nonstationary kernels (\eg Gibbs kernel). This flexibility allows \shortSpace to capture a wide range of behaviors, from globally stationary phenomena to cases where smoothness and variability change across the input domain. We highlight that \shortSpace can use $M$ kernels of the same type (e.g., $M$ Gaussians), with each component learning specific patterns in different parts of the domain.

The last key feature is that each term in Equation \ref{eq preactivation} leverages kernel closure Properties \ref{def scaling}$-$\ref{def warping} to ensure its validity. For instance, the terms $\wb^{\top}_m(\xb) \wb_m(\xb^{\prime})$ and $\bb^\top(\xb) \bb(\xb^{\prime})$ are valid kernels, as they result from applying Property \ref{def warping} to the linear kernel. The product $\wb^{\top}_m(\xb) \wb_m(\xb^{\prime}) c_m(\xb, \xb^{\prime})$ also ensures validity, according to Property \ref{def product}. More generally, by applying Properties \ref{def scaling}$-$\ref{def product}, it can be shown that the pre-activation $z(\xb,\xb^\prime)$ itself is a valid kernel. 

Choosing an appropriate activation function $\phi(\cdot)$ is essential to guarantee that the final result is a valid kernel. As discussed in Section \ref{subsec closure properties}, potential candidates for $\phi\parens{\cdot}$ include, but are not limited to, $\exp\parens{\cdot}$, $\sinh\parens{\cdot}$ and $\cosh\parens{\cdot}$. The third feature discussed above is in fact essential for ensuring the well-posedness of \textit{any} nonstationary kernel. Just as PointNet \cite{qi2017pointnet}, an NN architecture for processing point clouds, ensures permutation invariance at each of its building blocks, nonstationary kernels must guarantee kernel validity at every stage of their construction.

This combination of features enables \shortSpace to provide a flexible, customizable, and interpretable framework for kernel learning while ensuring kernel validity \textit{by construction}. In Section \ref{subsec illustrative example}, we demonstrate the explainability of the proposed kernel, and in Section \ref{subsec sensitivity}, we conduct sensitivity analyses to provide more insights on its key design choices.

\subsection{Explainability of \short: Illustrative Example} \label{subsec illustrative example}
We demonstrate the behavior of the proposed method on the \analyticI \ problem introduced in Section \ref{subsec benchmarks}. We use a zero-mean GP with \short, trained via MLE on the same $50$ data points as in \cite{noack2024unifying}. As base kernels, we employ Gaussian, periodic and Matérn kernels, \ie $M=3$, and use two independent neural networks---each with two hidden layers and four neurons per layer---to learn the weights $\braces{\wb_m(x; \betab_{\wb_m})}_{m=1}^{M}$ and bias $\bb(x; \betab_{\bb})$.

\begin{figure}[!h] 
    \centering
    \includegraphics[width=0.68\linewidth]{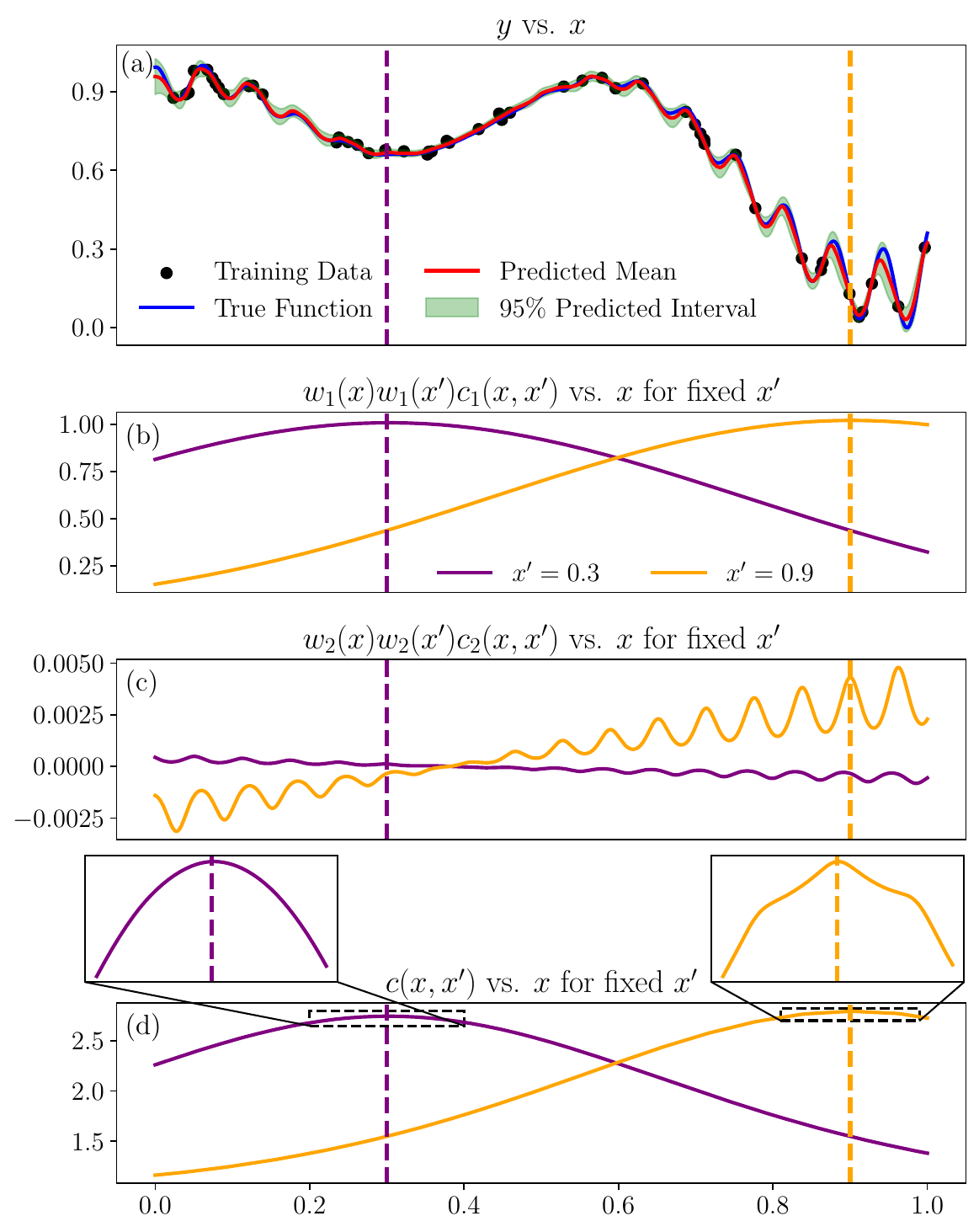}
    \caption{Prediction of a GP using \shortSpace on the \analyticI \ problem, along with two of the learned weighted base kernels and the resulting covariance function evaluated at two reference points.}
    \label{fig 1d illustrative example}
\end{figure}

The main results are illustrated in Figure \ref{fig 1d illustrative example}. Subplot (a) presents the predicted mean and $95\%$ confidence interval of the model. Subplots (b,c) show the learned weighted base covariances for the Gaussian and periodic kernels\footnote{For brevity, we omit the Matérn kernel and bias term in the subplots from Figure \ref{fig 1d illustrative example}.}, given by $\wb^{\top}_m(x) \wb_m(x^{\prime}) c_m(x, x^{\prime})$ for $m=1,2$. Subplot (d) illustrates the resulting covariance function $c(x,x')$ from \short. 
The functions in subplots (b-d) are evaluated at two reference points exhibiting different behaviors: smooth at $x^{\prime}=0.3$ (purple) and high-frequency variations at $x^{\prime}=0.9$ (orange).

From Figure \ref{fig 1d illustrative example}(a), we observe that our GP (1) produces highly accurate predictions, achieving an RMSE of $0.013$ which is notably smaller than $0.038$ reported in \cite{noack2024unifying}, and (2) provides well-calibrated prediction intervals, obtaining an NNOIS of $0.189$ (see Eq. \ref{eq nnois}). 
Figures \ref{fig 1d illustrative example}(b) and \ref{fig 1d illustrative example}(c) illustrate the learned weighted Gaussian and periodic kernels, respectively. The weighted Gaussian kernel exhibits a similar trend at both reference points where it has a maximum that is smoothly decreased. 
Unlike the Gaussian kernel, the weighted periodic kernel varies adaptively. Specifically, for $x^{\prime}=0.3$, the kernel exhibits little to no variation across the entire domain, aligning with the smooth nature of the underlying function. In contrast, at $x^{\prime}=0.9$, the kernel adjusts to local variations, \ie it exhibits an oscillating pattern only in regions where the underlying function does as well. Finally, Figure \ref{fig 1d illustrative example}(d) presents the resulting covariance function obtained by Eq. \ref{eq kernelX}. At a large scale, the covariance functions for both reference points appear similar. However, a closer look (see insets) reveals clear distinct behaviors: $c(x,x^\prime=0.9)$ exhibits a more oscillatory structure, driven by the learned weighted periodic kernel, whereas $c(x,x^\prime=0.3)$ presents a smoother behavior.

These results highlight the flexibility and interpretability of \shortSpace for kernel learning in GPs. By dynamically adjusting the contribution of different base kernels, the model effectively captures both smooth and high-frequency behaviors in different regions of the input space. This adaptive behavior is crucial for modeling complex, nonstationary functions where stationary kernels inherently fall short.
    \section{Results and Discussions} \label{sec results}

We begin this section by introducing the datasets and evaluation metrics. Then, we conduct sensitivity analyses to evaluate the dependence of our method on various design choices.
Finally, we present a series of comparative studies where we assess the performance of \shortSpace against other nonstationary kernels. The numerical experiments in this section are implemented using the open-source Python package \gp \ \cite{gp+}.

All simulations are repeated 16 times independently (except engineering problems in Section \ref{subsec engineering}, which are repeated 8 times) to ensure the results are representative.
 Also, all the GPs are trained via the L-BFGS optimizer with a learning rate of $0.01$. For further implementation and optimization details, please see Section \ref{sec implementation} in the Appendix.

\subsection{Benchmark problems and metrics} \label{subsec benchmarks}

We consider seven benchmark problems with varying levels of nonlinearity and dimensionality. These include analytic functions adopted from the literature \cite{noack2024unifying, picheny2013benchmark} and real-world datasets derived from engineering applications \cite{chen2024adaptive, deng2023data, camacho_elevators, delve_pumadyn32}. 

\textbf{\analyticI}: we consider the following 1D function \cite{noack2024unifying}:
\begin{equation}
\begin{aligned}
    & f(x)=\frac{1}{3.94}\left(\sin (5 x)+\cos (10 x)\right)  \\ 
    & + 1.435(x-0.4)^2 \cos (100 x)+0.659
\end{aligned}
\label{eq analytic 1D}
\end{equation}
where $x \in \brackets{0,1}$. We use $55$ points randomly drawn via Sobol sequence for our training dataset, and corrupt these samples with noise $\epsilon \sim \mathcal{N}(0, \lambda^2 = 10^{-4})$.

\textbf{\analyticII}: we design the following 1D function:
\begin{equation}
f(x)=
\begin{cases}
\operatorname{env}(x)+0.1\,\sin\left(8\pi x\right), & 0 \le x < 2, \\[1mm]
0.5\,e^{\,x-2}\,\sin\left(2\pi x\right), & 2 \le x < 4, \\[1mm]
\sin\Biggl( 2\pi \Bigl[2+(x-4)^2\Bigr]\,\frac{x-4}{2}\Biggr), & 4 \le x < 6, \\[1mm]
\displaystyle
\begin{cases}
16t, & 0 \le t < 0.25, \\[2mm]
8-16t, & 0.25 \le t < 0.5,
\end{cases}
& 6 \le x < 8, \\[1mm]
\sin\left(2\pi x\right)+0.5\,\sin\left(8\pi x\right), & 8 \le x \le 10,
\end{cases}
\end{equation}
where
\[
t = (x-6) \bmod 0.5,
\]
with $x \in \brackets{0,10}$ and 
\[
\operatorname{env}(x)=
\begin{cases}
4x, & 0 \le x < 0.5, \\[1mm]
2, & 0.5 \le x \le 1.5, \\[1mm]
4(2-x), & 1.5 < x \le 2.
\end{cases}
\]
We draw $140$ samples via Sobol sequence, and corrupt them with noise $\epsilon \sim \mathcal{N}(0, \lambda^2 = 10^{-4})$.




\textbf{\volcano}: we consider a dataset containing the terrain elevation of Mount Saint Helens as a function of planar spatial location \cite{chen2024adaptive}. This terrain exhibits nonstationary behavior due to the presence of prominent environmental features, which contribute to spatially varying smoothness and complexity of the terrain. We use a training dataset consisting of $400$ data points drawn via Sobol sequence, and corrupt them with noise $\epsilon \sim \mathcal{N}(0, \lambda^2 = 1)$.


\textbf{\hartmann}: we consider the 6D Hartmann function \cite{dixon1978global, picheny2013benchmark}:
\begin{equation}
f(\xb) \;=\; - \sum_{i=1}^{4} \alpha_i \exp \Biggl(-\sum_{j=1}^{6} A_{ij}\Bigl(x_j - P_{ij}\Bigr)^2\Biggr),
\end{equation}
where $\xb \in [0,1]^6$ and the constants \(\alpha_i\), \(A_{ij}\), and \(P_{ij}\) are typically set as follows:
\[
\boldsymbol{\alpha} = [1.0,\; 1.2,\; 3.0,\; 3.2],
\]
\[
A = 
\begin{bmatrix}
10 & 3 & 17 & 3.5 & 1.7 & 8 \\
0.05 & 10 & 17 & 0.1 & 8 & 14 \\
3 & 3.5 & 1.7 & 10 & 17 & 8 \\
17 & 8 & 0.05 & 10 & 0.1 & 14
\end{bmatrix},
\]
\[
P = 10^{-4} \times
\begin{bmatrix}
1312 & 1696 & 5569 & 124  & 8283 & 5886 \\
2329 & 4135 & 8307 & 3736 & 1004 & 9991 \\
2348 & 1451 & 3522 & 2883 & 3047 & 6650 \\
4047 & 8828 & 8732 & 5743 & 1091 & 381
\end{bmatrix}.
\]

This function has multiple local minima and is inherently nonstationary where small input perturbations can induce large variations in the output. We use $800$ samples for training drawn via Sobol sequence, and corrupt them with noise $\epsilon \sim \mathcal{N}(0, \lambda^2 = 10^{-4})$.

\textbf{\rom}:
we consider our in-house dataset derived from a reduced‑order model for multiscale fracture simulations of cast aluminum alloys \cite{deng2023data}. Each sample contains $6$ real-valued features (four microstructural descriptors and two material properties) and a single output corresponding to the toughness of the microstructure. In each repetition, we train the models with $200$ samples randomly drawn from the full dataset and test their performance on the remaining $50$ samples.

\textbf{\elevators}:
this dataset \cite{camacho_elevators} captures the behavior of an aircraft elevator control system. It has $18$ continuous inputs representing the aircraft’s current flight state and recent elevator commands, and a single response showing the desired elevator deflection for the next control step. We discard eight of the input features as they do not affect the response and for each repetition we randomly draw $1{,}750$ training samples ($20\%$ of the full training set) and evaluate the performance of the models on $7{,}847$ test samples.

\textbf{\pumadyn}: this dataset \cite{delve_pumadyn32} was derived from a realistic high-fidelity simulation of the dynamics of a Unimation PUMA 560 robot arm. We use the \textit{puma32H} variant with moderate output noise, where each data point contains $32$ continuous input features including angular positions, velocities, torques and additional dynamic parameters of the robot arms, and a single continuous target corresponding to the angular acceleration of the arm’s sixth link.  For each repetition, we randomly draw $2{,}699$ training samples ($60\%$ of the full training set) and use $3{,}692$ test samples to assess the performance of the models.

For all our experiments, we scale both inputs and output using the mean and standard deviation computed from the training set (to avoid data leakage). The same transformation is then applied to the test set for consistency.
We evaluate the performance of all models across all examples using a noiseless test set consisting of $N_{test}$ samples. Our evaluation metrics are the normalized root mean squared error (NRMSE) and the normalized negatively oriented interval score (NNOIS):
\begin{subequations} 
    \begin{equation} 
        \text{NRMSE} = \frac{1}{s} \sqrt{\frac{1}{N_{test}}\sum_{i=1}^{N_{test}}\parens{\bar{m}(\xb_{i}) - f(\xb_{i})}^2},
    \label{eq nrmse}
    \end{equation}
    \begin{equation}
        \begin{aligned}
            \text{NNOIS} &= \frac{1}{s} \frac{1}{N_{test}} \sum_{i=1}^{N_{test}}\parens{\bar{u}(\xb_{i}) - \bar{l}(\xb_{i})} \\ & + \frac{2}{\alpha}\parens{\bar{l}(\xb_{i}) - f(\xb_{i})} \mathbbm{1}\braces{f(\xb_{i}) < \bar{l}(\xb_{i})} \\
            & + \frac{2}{\alpha}\parens{f(\xb_{i}) - \bar{u}(\xb_{i})} \mathbbm{1}\braces{f(\xb_{i}) > \bar{u}(\xb_{i})},
        \end{aligned}
        \label{eq nnois}
        \end{equation}
    \label{eq metrics}
\end{subequations}   
where $s$ is the output standard deviation of the $N_{test}$ test samples, and $\bar{l}(\xb_{i})$ and $\bar{u}(\xb_{i})$ denote the predicted lower and upper bounds of the prediction interval for the $i$-th test sample, respectively. We employ $95\%$ prediction intervals, \ie $\alpha=0.05$, and thus its endpoints can be computed via $\bar{l}(\xb_{i}) = \bar{m}(\xb_{i}) - 1.96 \bar{c}\parens{\xb_{i},\xb_{i}}$ and $\bar{u}(\xb_{i}) = \bar{m}(\xb_{i}) + 1.96 \bar{c}\parens{\xb_{i},\xb_{i}}$. $\mathbbm{1}\{\cdot\}$ is an indicator function which is $1$ if its condition holds and zero otherwise. 
For both metrics in Eq. \ref{eq metrics} lower values are better, with NRMSE reflecting the accuracy of the mean predictions and NNOIS accounting for the quality of the prediction intervals.

\begin{figure*}[!t] 
    \centering
    \includegraphics[width=\linewidth]{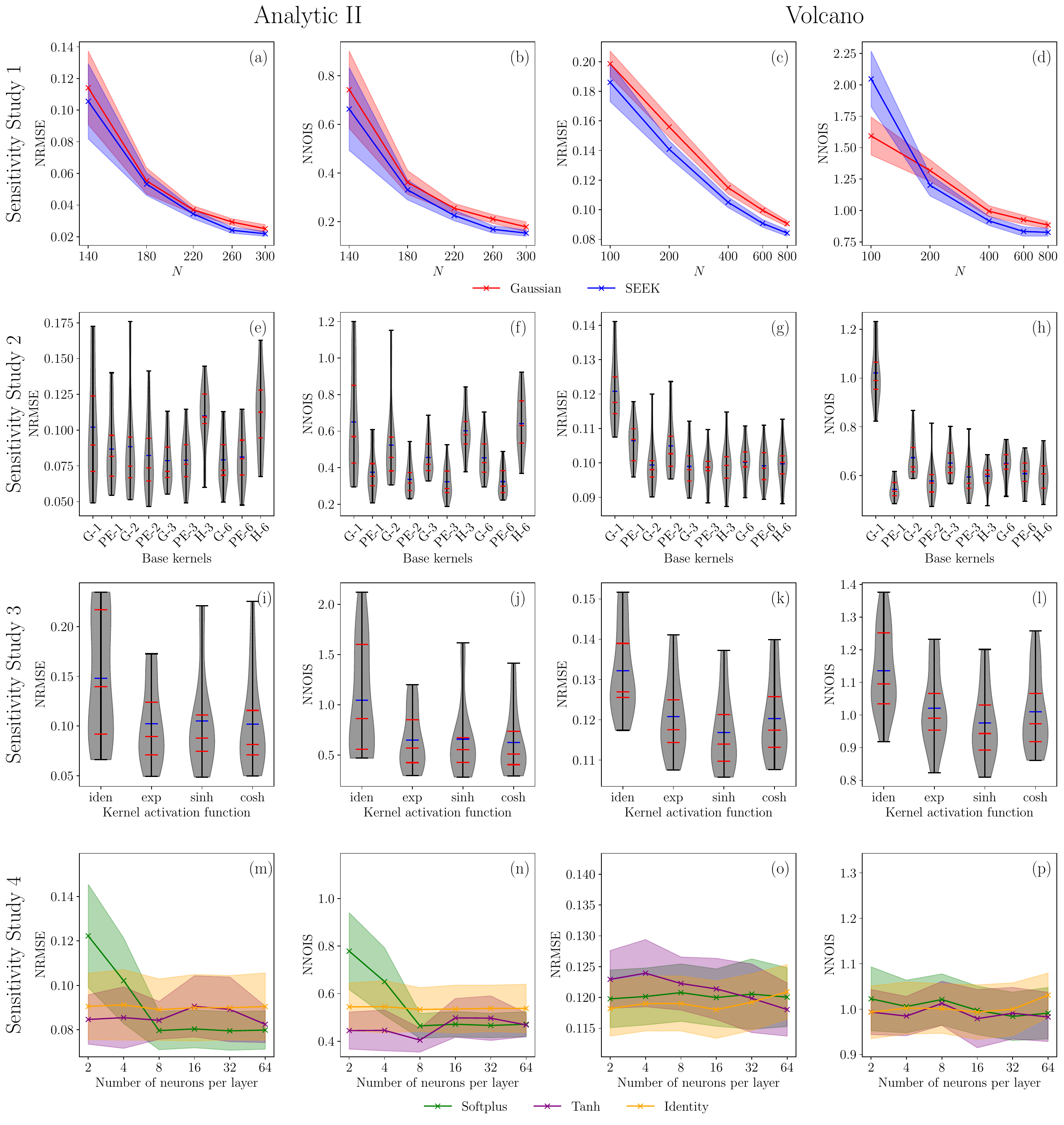}
    \caption{Sensitivity studies: We conduct four sensitivity studies on two benchmark problems to evaluate the impact of key design choices on the performance of \short. The results are based on 16 repetitions.}
    \label{fig sensitivity}
\end{figure*}

\subsection{Sensitivity Studies} \label{subsec sensitivity}
We conduct four sensitivity studies to assess the impact of key design choices on the performance of \short:
\begin{itemize}
    \item \textbf{Sensitivity Study 1 (Convergence behavior)}: we examine the convergence behavior as a function of data size $N$. For \short, we use a single Gaussian kernel as the base kernel and compare it to the Gaussian kernel in Equation \ref{eq Gaussian kernel}.

    \item \textbf{Sensitivity Study 2 (Kernel structure)}: we analyze the influence of the number and types of base kernels used within \short. In the plots, we use the format ``letter-number'' to indicate the type and total number of base kernels. Specifically, ``G'' refers to the Gaussian kernel,``PE'' to the power exponential kernel, and ``H'' to a hybrid kernel (a combination of Gaussian, periodic, and Matérn kernels).

    \item \textbf{Sensitivity Study 3 (Kernel activation function)}: we evaluate the impact of $\phi$ by testing exponential, hyperbolic sine, hyperbolic cosine, and identity (denoted as ``iden'') activation functions.

    \item \textbf{Sensitivity Study 4 (Network architecture)}: we investigate the impact of the number of neurons and the choice of activation function in the NNs\footnote{Only two hidden layers are used.} employed for learning $\wb$ and $\bb$. We test three different activation functions: softplus, tanh, and identity (\ie no activation).
    
\end{itemize}  

We evaluate the performance of the models in terms of NRMSE and NNOIS. The results of these studies are summarized in Figure \ref{fig sensitivity}, where each row corresponds to a different study. The analyses are assessed on the \analyticII\ function and the \volcano\ dataset. We note that the solid lines and shaded bands in panels (a-d) and (m-p) represent the mean and $95\%$ confidence intervals across repetitions, respectively. The red lines in panels (e-h) and (i-l) indicate the first, second and third quartiles across repetitions, while the blue line indicates the mean.

From Sensitivity Study 1 in Figure \ref{fig sensitivity}(a-d), we observe that \shortSpace achieves lower NRMSE and NIS compared to the Gaussian kernel in both benchmark problems for all sample sizes, except for $N=100$ in the \volcano\ dataset. This discrepancy is attributed to the fact that $100$ samples are insufficient to capture meaningful patterns about the underlying function, leading to both models clearly underfitting the data. In addition, as the number of samples increases, the performance gap between \shortSpace and the Gaussian kernel decreases. This behavior is expected due to the interpolation capabilities inherent to GPs. More specifically, from Equations \ref{eq posterior mean}  and \ref{eq posterior cov}, and assuming noiseless samples, we observe that, when making predictions on the training dataset $\Xb$:
\begin{subequations} 
    \begin{equation*} 
        \bar{m}\parens{\Xb} = m\parens{\Xb} + \cancelto{\boldsymbol{I}}{c(\Xb,\Xb) \Cb^{-1}}\parens{\yb - \mb} = \yb,
    \end{equation*}
    \begin{equation*}
        \bar{c}\parens{\Xb,\Xb} = c\parens{\Xb,\Xb} - \cancelto{\boldsymbol{I}}{c(\Xb,\Xb) \Cb^{-1}}c\parens{\Xb,\Xb} = \boldsymbol{0}.
    \end{equation*}
\end{subequations}

These results highlight the importance of nonstationary kernels in low-to-mid data regimes, where they significantly influence the model’s predictions. However, in high data regimes, the inherent interpolation capabilities of GPs dictate the predictions, making the choice of kernel less relevant.

This observation poses an interesting dilemma: while nonstationary kernels are more beneficial in low-to-mid data regimes, they typically rely on a larger number of parameters. This increased flexibility makes them more prone to overfitting compared to stationary kernels, especially when samples are scarce. This trade-off implies that one must constrain the kernel’s complexity (\ie number of parameters) according to both the input dimensionality and the available dataset size to prevent overfitting. 


From Sensitivity Study 2 in Figure \ref{fig sensitivity}(e-h), we observe that \shortSpace benefits significantly from an increased number of base kernels in both problems. For example, the model performs noticeably better when using 6 Gaussian kernels (G-6) compared to just 1 (G-1). However, the model does not show substantial improvement from combining different types of kernels when a high number of base kernels is used, as seen in the comparison between models using only Gaussian or power exponential kernels versus the hybrid kernel.

\begin{figure*}[!t] 
    \centering
    \begin{subfigure}{\linewidth}
        \includegraphics[width=\linewidth]{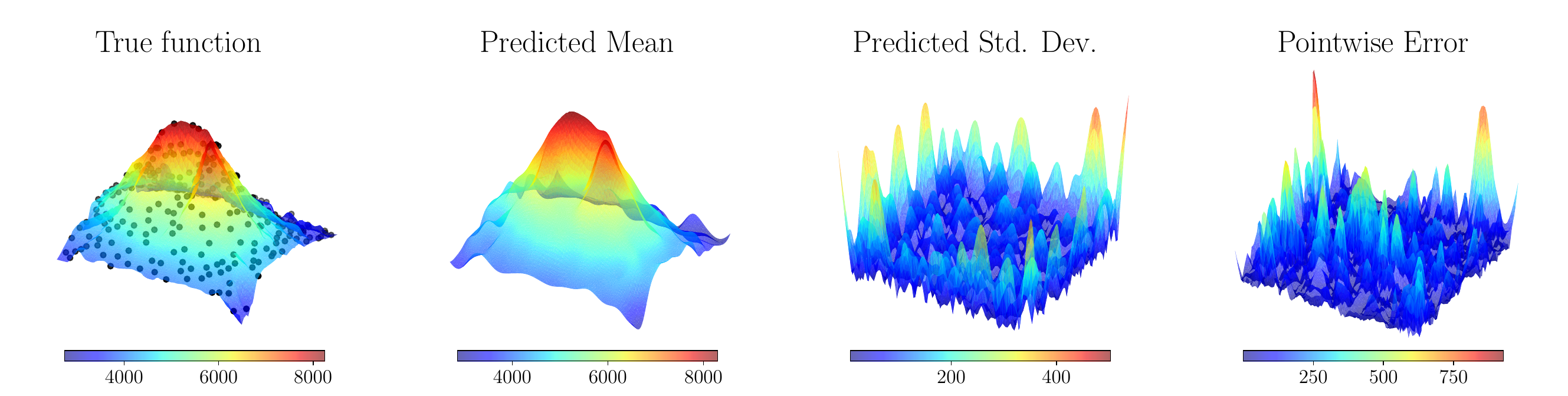}
        \caption{1 Gaussian as base kernel.}
    \end{subfigure}
    \vfill
    \begin{subfigure}{\linewidth}
        \includegraphics[width=\linewidth]{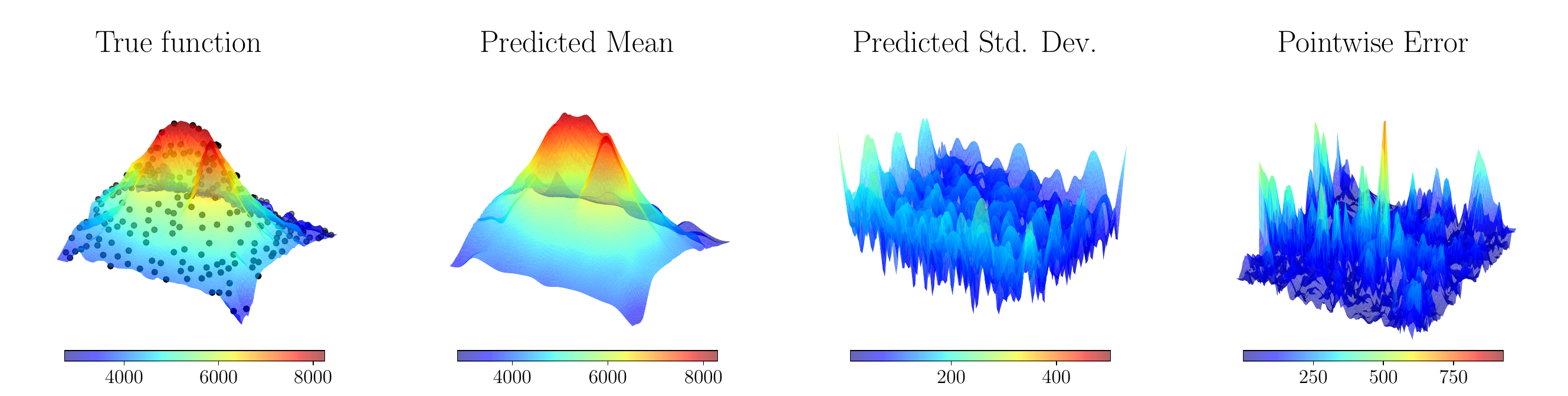}
        \caption{6 Gaussians as base kernels.}
    \end{subfigure}
    \caption{Our proposed kernel \shortSpace with (a) 1 and (b) 6 Gaussian kernels as base kernels on the \volcano \ dataset: Increasing the number of base kernels reduces error and improves confidence, as reflected in the colorbars (same ranges are used in (a) and (b) to facilitate direct comparison). Black dots represent the training data points.}
    \label{fig sensitivity study two}
\end{figure*}

To visualize the above behavior, we present the predictions of a GP with \shortSpace using 1 and 6 Gaussian kernels as the base kernels, see Figure \ref{fig sensitivity study two}. While the model's performance with a single Gaussian kernel in Figure \ref{fig sensitivity study two}(a) is already strong, the use of six Gaussian kernels in Figure \ref{fig sensitivity study two}(b) provides enhanced flexibility, allowing the model to learn more complex patterns. This results in a noticeable reduction in overall error and more confident predictions.

From Sensitivity Study 3 in Figure \ref{fig sensitivity}(i-l), we observe that the model performs better in terms of both NRMSE and NNOIS if $\phi(\cdot)$ has nonlinearity.  This is consistent with the discussion in Section \ref{subsec method proposed framework}, given that the activation function enhances the flexibility of the kernel by introducing additional interactions between the weighted base kernels. Furthermore, we do not observe a significant difference in performance as $\phi(\cdot)$ switches between the exponential, hyperbolic sine, or hyperbolic cosine activation functions. This is also expected, as these functions are quite similar.

From Sensitivity Study 4 in Figures \ref{fig sensitivity}(m-p), we observe that the model performs well even $\wb$ and $\bb$ are based on the identity activation function, suggesting that a simple linear mapping (instead of an NN) could provide decent accuracy for learning these functions (note that the model does benefit from nonlinear activation functions such as tanh or softplus but this benefit is not substantial in these examples). Once again, the model exhibits strong robustness against overparameterization since its performance remains consistent as the number of neurons is increased.

\begin{figure*}[!t] 
    \centering
    \includegraphics[width=\linewidth]{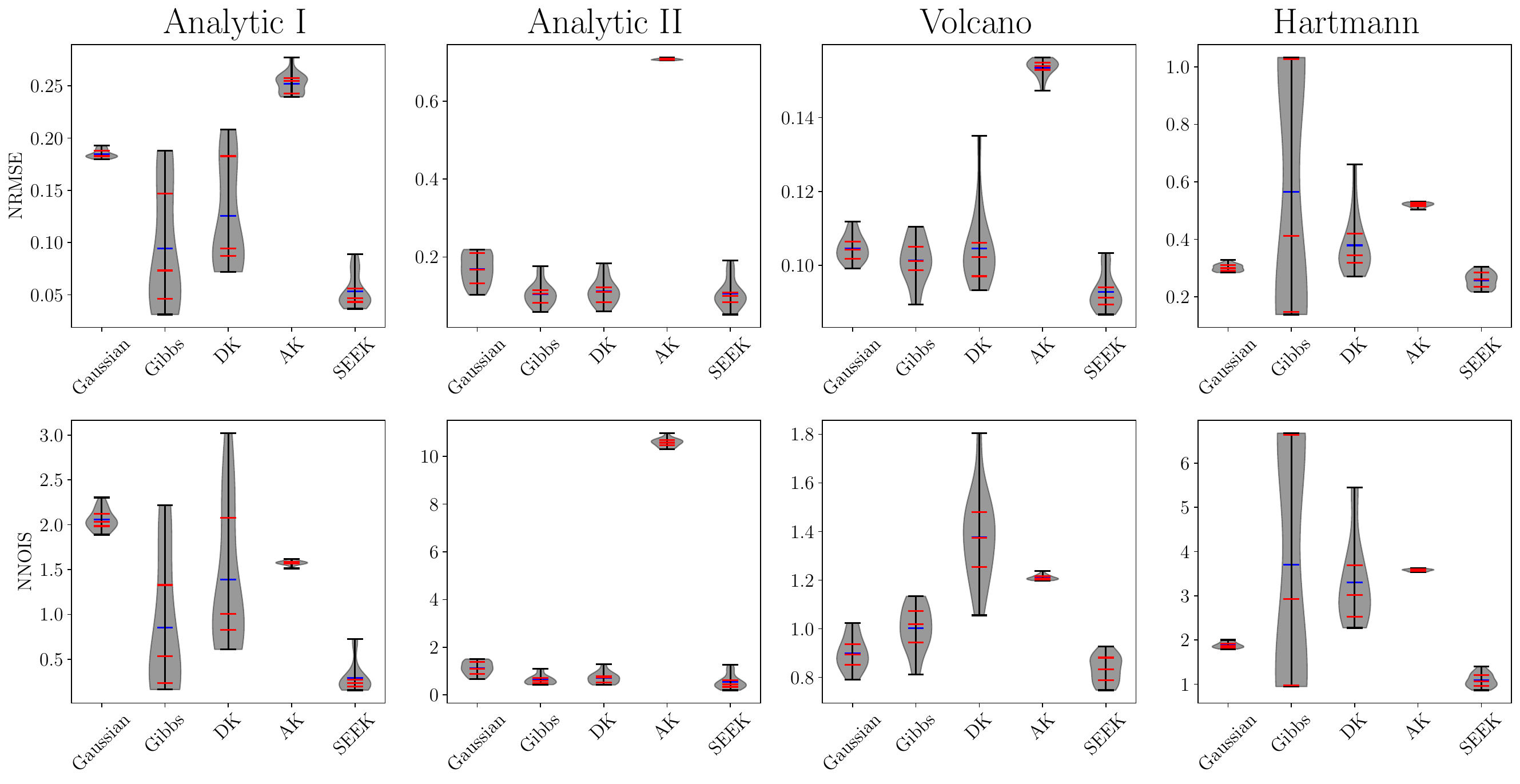}
    \caption{Comparative studies: We compare \shortSpace against other nonstationary kernels on four benchmark problems with varying degrees of nonlinearity and dimensionality.}
    \label{fig comparative}
\end{figure*}

\subsection{Comparative Studies} \label{subsec comparative}
We compare \shortSpace with one Gaussian kernel as the base kernel against four stationary/nonstationary kernels: Gaussian (Eq. \ref{eq Gaussian kernel}), Gibbs \cite{gibbs1998bayesian}, deep kernel (DK) \cite{wilson2016deep}, and attentive kernel (AK) \cite{chen2024adaptive}. 

To ensure a fair comparison, we use two hidden layers for the NNs used in the nonstationary kernels, and ensure that the number of learnable parameters remains comparable across different approaches. All hidden layers employ the softplus activation function, while all output layers use a linear (identity) activation. More specifically, in \shortSpace we use a single Gaussian as the base kernel, and use two NNs with two hidden layers each having $2P$ neurons to model $\wb(\xb)$ and $\bb(\xb)$, where $P$ is the input dimensionality. The output layer of $\wb$ contains a single neuron, while $\bb$ has two output neurons. The NN in the deep kernel consists of two hidden layers with $4P$ neurons each, followed by a linear output layer of dimension $P$. Similarly, the NN parameterizing the lengthscales in the Gibbs kernel comprises two hidden layers of size $4P$ and a linear output layer with $P$ neurons. Lastly, our implementation of the attentive kernel employs $10$ Gaussian kernels with fixed, equally spaced lengthscale parameters. The two NNs (denoted as the $z$- and $w$-networks in \cite{chen2024adaptive}) used to compute similarity and visibility attention scores have been unified, following the authors' recommendations, and designed so that the number of learnable parameters is on par with the other nonstationary kernels. 

The results of the comparative studies are summarized in Figure \ref{fig comparative}.
For the \analyticI \ problem, \shortSpace achieves the best performance in terms of NRMSE and NNOIS across all kernels. This is a remarkable result, especially considering that, as shown in Figures \ref{fig sensitivity}(e-h), \shortSpace achieves even better performance if provided with a richer set of base kernels. However, as mentioned earlier, we intentionally restricted the model to a single base kernel to maintain a comparable number of learnable parameters. Other nonstationary kernels, such as the Gibbs and deep kernels, also show noticeable improvements over the Gaussian kernel. However, they do not demonstrate the same level of robustness across different repetitions compared to \short, as reflected by their wider variations in the violin plots. 

In the \analyticII \ and \volcano \ problems, a different trend is observed: although \shortSpace remains the top-performing kernel, the stationary Gaussian kernel performs comparably or even better than the other nonstationary kernels. We attribute this to the higher density of training data in these two problems, which reduces the benefits of modeling nonstationarity by increasing kernel's flexibility. This aligns with the expected behavior of GPs, as discussed in Section \ref{subsec sensitivity}: in higher data regimes, the natural interpolation properties of GPs, combined with the smooth prior induced by the Gaussian kernel, can yield sufficiently strong predictive performance.

In the \hartmann \ problem, we observe that \shortSpace outperforms the Gaussian, deep and attentive kernels in terms of NRMSE and NNOIS. However, the Gibbs kernel occasionally achieves lower metric values (but also higher ones), as indicated by its wider violin plots. 

In our studies, we noticed that more complex nonstationary kernels (with higher number of parameters) generally require more data to outperform the Gaussian kernel in high-dimensional, sparse settings where effectively learning input-dependent variations becomes especially challenging. We attribute this to the fact that, while these nonstationary kernels introduce additional flexibility to capture such local variations, they also incur a higher risk of overfitting when data are limited. In contrast, the Gaussian kernel imposes a globally smooth prior, acting as an effective regularizer. This observation aligns with prior works \cite{remes2018neural, risser2020bayesian, noack2023exact}, which combine flexible nonstationary kernels with sparse approximation techniques to allow the GP to leverage large datasets in high dimensions. 


Finally, we would like to comment on the attentive kernel \cite{chen2024adaptive}. As shown in Figure \ref{fig comparative}, we found that it does not perform well on any of the benchmark problems. While we have carefully validated our implementation against the authors' provided code in \cite{chen2024adaptive} to ensure consistency, the observed performance in our studies may be caused due to variations in experimental settings or hyperparameter choices. In addition, we suspect that a key factor contributing to its inefficiency is that the lengthscale range recommended by the authors may not generalize effectively across problems with varying dimensionality and complexity.

\subsection{Engineering Problems} \label{subsec engineering}
In this section, we evaluate \shortSpace on three challenging, high-dimensional engineering regression problems (described in Section \ref{subsec benchmarks}): our in‑house \rom \ dataset, and the popular \elevators \ and \pumadyn \ benchmarks. For these three problems, we compare \shortSpace against the standard Gaussian kernel. The \elevators \ and \pumadyn \ datasets have been widely tested in the literature, which allows us to also compare \short's performance against regression models other than GPs \cite{moura2021mine, zhou2023synthetic, sigrist2021ktboost, waxman2024dynamic,lay2012artificial}. These studies aim to show that \shortSpace provides more accurate mean predictions and improved uncertainty quantification in realistic engineering datasets.

\begin{figure*}[!b]
    \centering
    \includegraphics[width=0.65\linewidth]{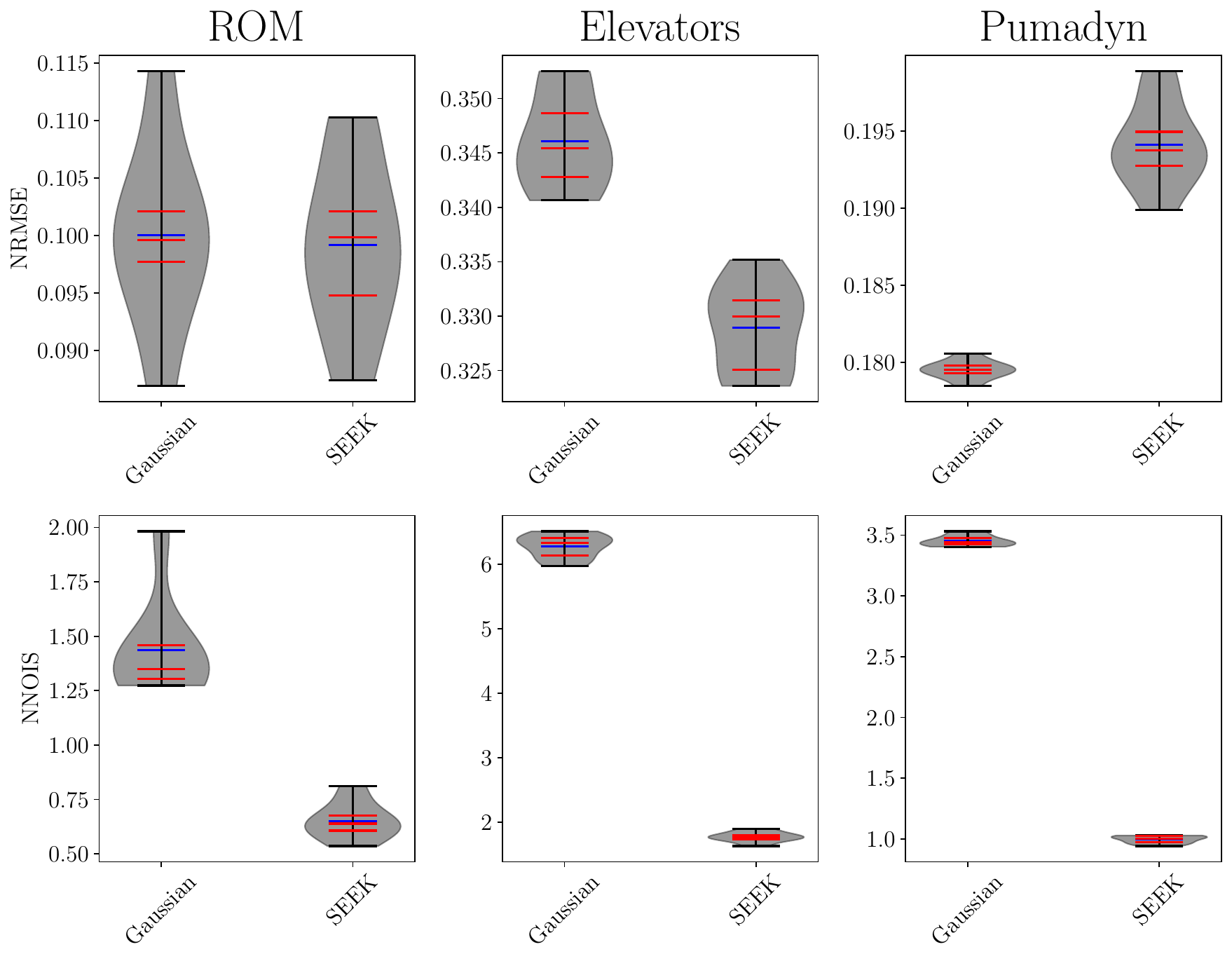}
  \caption{Engineering Problems: We compare the performance of GP using \shortSpace versus a standard Gaussian kernel on three high-dimensional engineering problems.}
  \label{fig:engProblems-metrics}
\end{figure*}


Similar to Section \ref{subsec comparative}, we parameterize the weight function $\wb(\xb)$ and bias function $\bb(\xb)$ with two separate feed-forward NNs.
Each network comprises two hidden layers, with (4, 2) neurons for \rom, (8, 4) for \elevators, and (16, 8) for \pumadyn.
We chose these layer widths to roughly scale with input dimensionality while keeping the total parameter count reasonable and manageable.

Figure \ref{fig:engProblems-metrics} summarizes the results where we observe that in the \rom \ dataset \shortSpace achieves roughly a two-times improvement in terms of NRMSE and NNOIS across repetitions. In the \elevators \ dataset, \shortSpace also consistently outperforms the Gaussian kernel, achieving a marginally lower NRMSE while reducing the NNOIS by over three times. However, we observe a different trend in the \pumadyn \ dataset: the single-Gaussian kernel yields an slightly lower NRMSE, but \shortSpace still delivers an approximately three-fold improvement in terms of NNOIS. These results highlight, once again, \short's capability in providing more reliable and informative uncertainty estimates, which is particularly valuable in engineering design applications where trustworthy UQ is critical.

Finally, we highlight that across all our experiments we have found that GP models incorporating \shortSpace tend to converge more consistently and in fewer iterations than those using conventional kernels. In particular, on the \pumadyn \  problem, we observed that the single-Gaussian kernel failed to converge under the L-BFGS optimizer and only produced the results shown in Figure \ref{fig:engProblems-metrics} when trained with AdamW at an unusually high learning rate of $1.0$ to escape local minima. In contrast, GPs with \shortSpace converged smoothly under the standard optimizer settings employed throughout our studies.  This observation holds across a variety of datasets and kernel settings, suggesting that reasonably increased kernel flexibility can not only boost predictive performance but also improve optimization stability and efficiency.

    \section{Conclusions} \label{sec conclusion}
We introduce \short, a flexible, customizable, and interpretable framework for kernel learning in GPs. We believe this method offers a novel and systematic approach for designing nonstationary kernels while guaranteeing kernel validity. Our sensitivity analyses highlight that GPs that use \short~are very robust against overparameterization. In addition, our extensive comparative studies demonstrate that \shortSpace outperforms other stationary/nonstationary kernels in both prediction accuracy and uncertainty quantification. 


A natural direction for extending \shortSpace is to integrate it with sparse approximation techniques to enhance its scalability to big data applications. In addition, as we discussed in Section \ref{sec method}, \shortSpace is inspired by artificial neurons in NNs. This suggests a promising direction: exploring its performance when stacked layer by layer, similar to how neurons are combined in NN architectures. Another potential extension is to make the activation function itself learnable, which could further boost the model’s performance. Finally, applying \shortSpace to real-world problems that require reliable uncertainty quantification, such as Bayesian optimization, also presents an exciting avenue for future work.

\section{Acknowledgments}
We appreciate the support from the Office of Naval Research (grant number $N000142312485$) and the National Science Foundation (grant numbers $2238038$ and $2525731$).
    \appendix
\section{Implementation And Optimization Details} \label{sec implementation}
All experiments in Sections \ref{subsec sensitivity} and \ref{subsec comparative}  have been implemented using the Python package GP+ \cite{gp+}. For the optimization of the models, we used the L‑BFGS optimizer from PyTorch with a fixed learning rate of $0.01$. To reduce the risk of converging to suboptimal solutions, we used the strategy of rerunning the optimization of models with different initializations of the learnable parameters. This is a common and well-known strategy, especially for training models like GPs. Although it is not necessary to use this many reruns, to give each model enough opportunity to show its best performance and reduce the effect of the initial values of learnable parameters on the performance of the models, we used $80$ reruns.

In practice, we don't need to reinitialize these many times to get decent results. Yet, it should be mentioned that it makes more sense to increase the number of initializations with the increase of the problem's dimensionality, as the optimization space enlarges.

Each model was trained for a maximum of $2{,}000$ epochs while we applied early stopping if the computed loss failed to improve for $20$ consecutive epochs. All experiments were conducted on a machine equipped with an Intel Core i9‑14900KF CPU and 64 GB of RAM.
    \printbibliography
\end{document}